\documentclass{article}

     \PassOptionsToPackage{numbers, compress}{natbib}

    \usepackage[preprint]{neurips_2024}

\usepackage{bbm}
\usepackage[utf8]{inputenc} 
\usepackage[T1]{fontenc}    
\usepackage{hyperref}       
\usepackage{url}            
\usepackage{booktabs}       
\usepackage{amsfonts}       
\usepackage{nicefrac}       
\usepackage{microtype}      
\usepackage{xcolor}         

\usepackage{amsmath}
\usepackage{graphicx}
\usepackage{wrapfig}
\usepackage{algorithm}
\usepackage{amssymb}
\usepackage{multicol,multirow,subcaption}
\newtheorem{lemma}{Lemma}
\newtheorem{theorem}{Theorem}
\newcommand{\expnum}[2]{{#1}\mathrm{e}{#2}}

\title{Fast Deep Predictive Coding Networks for Videos Feature Extraction without Labels}

\author{%
  Wenqian ~Xue, Chi Ding, Jose Principe \\
  Department of Electrical \& Computer Engineering\\
  University of Florida\\
  Gainesville, FL 32611 \\
  \texttt{w.xue@ufl.edu, ding.chi@ufl.edu, principe@cnel.ufl.edu}\\ 
  }

\begin{document}

\maketitle

\begin{abstract}
Brain-inspired deep predictive coding networks (DPCNs) effectively model and capture video features through a bi-directional information flow, even without labels. They are based on an overcomplete description of video scenes, and one of the bottlenecks has been the lack of effective sparsification techniques to find discriminative and robust dictionaries. FISTA has been the best alternative. This paper proposes a DPCN with a fast inference of internal model variables (states and causes) that achieves high sparsity and accuracy of feature clustering. The proposed unsupervised learning procedure, inspired by adaptive dynamic programming with a majorization-minimization framework, and its convergence are rigorously analyzed. Experiments in the data sets CIFAR-10, Super Mario Bros video game, and Coil-100 validate the approach, which outperforms previous versions of DPCNs on learning rate, sparsity ratio, and feature clustering accuracy. Because of DCPN's solid foundation and explainability, this advance opens the door for general applications in object recognition in video without labels. 
\end{abstract}

\section{Introduction}

Sparse model is significant for the systems with a plethora of parameters and variables, as it selectively activates only a small subset of the variables or coefficients while maintaining representation accuracy and computational efficiency. This not only efficiently reduces the demand and storage for data to represent a dynamic system but also leads to more concise and easier access to the contained information in the areas including control, signal processing, sensory compression, etc.

In the control theory sense, a model for a dynamic process is often described by the equations 
\begin{displaymath} 
\left\{ \begin{array}{l} {y_t} = {G_t}(x_t) + n_t\\ x_t= {F_t}(x_{t-1},u_{t})+w_t \end{array} \right. \end{displaymath}
where $y_t$ is a set of measurements associated with a changing state $x_t$ through a mapping function $G_t$, the states $x_t$, also known as the signal of interest, is produced from a past one $x_{t-1}$ and an input $u_{t}$ through an evolution function $F_t$,  $w_t$ is the measurement noise and $n_t$ is the modeling error. Given measurements $y_t$ and input $u_{t}$, the Kalman filter \cite{kalman0,kalman} has emerged as a widely-employed technique for estimating states \cite{kalman2,kalman3} and mapping functions using neural networks \cite{kalman4} in a sparse way \cite{sparseK1,sparseK2,sparseK3}. Therein, it is typically constrained to estimate one variable, namely the state. \textbf{Can both state and input variables be estimated?} For many dynamic plants characterized by natural and complex signals, latent variables often exhibit residual dependencies as well as non-stationary statistical properties. \textbf{Can data with non-stationary statistics be well represented?}  Additionally, (deep) neural networks (NNs) \cite{NN1,NN2,NN3,kalman4} with multi-layered structures are extensively used for sparse modeling of dynamic systems \cite{sparseNN1,sparseNN2,sparseNN3}. 
Similarly structured, convolutional NNs have demonstrated significant success in tasks such as target detection and feature classification in computer vision and control applications \cite{cnn1,cnn2,cnn3,cnn4,cnn5}. 
As we all know, these methods are mathematically uninterpretable, and the NN architecture is a feedforward pass through stacks of convolutional layers. As studied in \cite{hira}, a bi-directional information pathway, including not only a feedforward but also a feedforward and recurrent passing, is used by brain for effective visual perception. \textbf{Can dynamics be represented in an interpretable way with bi-directional connections and interactions?}

These goals can be achieved by the hierarchical predictive coding networks \cite{pc1,pc2,pc3,pc4}, also known as deep-predictive-coding networks (DPCNs) \cite{dpcn3,dpcn4,dpcn6,dpcn1,dpcn7,dpcn5,dpcn}, where, inspired by \cite{hira}, a hierarchical generative model is formulated as \[\left\{ \begin{array}{l} {y_t^{l}} = {G_t}(x_t^{l}) + n_t^{l}\\ x_t^{l}= {F_t}(x_{t-1}^{l},u_t^{l})+w_t^{l} \end{array} \right.\] where $l $ denotes layers. Measurements for layer $l$ are the causes of the lower layer, i.e., $y_t^l=u_t^{(l-1)}$ for $l > 1$. The causes link the layers, and the states link the dynamics over time $t$. The model admits a bi-directional information flow \cite{brain1,dpcn5}, including feedforward, feedback, and recurrent connections. That is, measurements travel through a bottom-up pathway from lower to higher visual areas (for rapid object recognition) and simultaneously a top-down pathway running in the opposite direction (to enhance the recognition) \cite{brain2}. The previous DPCNs either use linear filters for sound \cite{dpcn3,dpcn4} or convolutions to better preserve neighborhoods in images \cite{dpcn6,dpcn1}. With fovea vision, non-convolutional DPCNs may offer a more automated and straightforward implementation \cite{dpcn,dpcn5}. In both types of DPCNs, the proximal gradient descent methods, such as fast iterative shrinkage-thresholding algorithm (FISTA) \cite{FISTA}, are frequently used for variable and model inferences in \cite{dpcn6,dpcn,dpcn5} for accelerated inference. \textbf{Can the DPCNs inference be faster while maintaining high sparsity?}

This paper answers these questions by studying vector DPCN with an improved inference procedure for both variable and models (dictionary) that is applicable to the two types, and that will be tested for proof of concept to model and capture objects in videos. Given measurements from the real world, the DPCNs infer model parameters and variables through feedforward, feedback, and recurrent connections represented by optimization problems with sparsity penalties. Inspired by the maximization minimization (MM) \cite{mm1} and the value iteration of reinforcement learning (RL) \cite{adp}, this paper proposes a MM-based unsupervised learning procedure to enhance the inference of DPCNs by introducing a majorizer of the sparsity penalty. This is called MM-DPCNs and offers the following advantages:
\begin{itemize}

\item The learning procedure does not need labels and offers accelerated inference.

\item The inference results guarantee sparsity of variables and representation accuracy of features.

\item Rigorous proofs show convergence and interpretability. 

\item Experiments validate the higher performance of MM-DPCNs versus previous DPCNs on learning rate, sparsity ratio, and feature clustering accuracy.

\end{itemize}


\section{Dynamic Networks for DPCNs}

\begin{wraptable}{r}{8cm}
\centering
    \caption{Detonations.}
    \begin{tabular}{cccccccccccccc}
    \toprule\midrule
    \multirow{1}{*}{$y_{t,n}^1$} & \multicolumn{1}{c}{$n$-th patch of video frame at time $t$}\\
    \midrule
    $y_{t,n}^l$, $l>1$ & {the causes from layer $l-1$}\\
    \midrule
    $x_{t,n}^l$ & state at layer $l$ for $y_{t,n}$\\
    \midrule
    $u_{t}^l$ & cause at layer $l$ for a group of $x_{t,n}^l$ \\
    \midrule
    $A^l, B^l, C^l$ & model parameters at layer $l$ \\
    \midrule\bottomrule
    \end{tabular}
    \label{tab:denotations}
\vspace{-1.5em}
\end{wraptable}
Based on the hierarchical generative model \cite{hira,dpcn} briefly reviewed in the Introduction, we now review the dynamic networks for DPCNs \cite{dpcn,dpcn5} in terms of sparse optimization problems for sparse model and feature extraction of videos. 

\begin{wrapfigure}{1}[0cm]{0pt}
\vspace{0cm}
\includegraphics[width=0.45\textwidth]{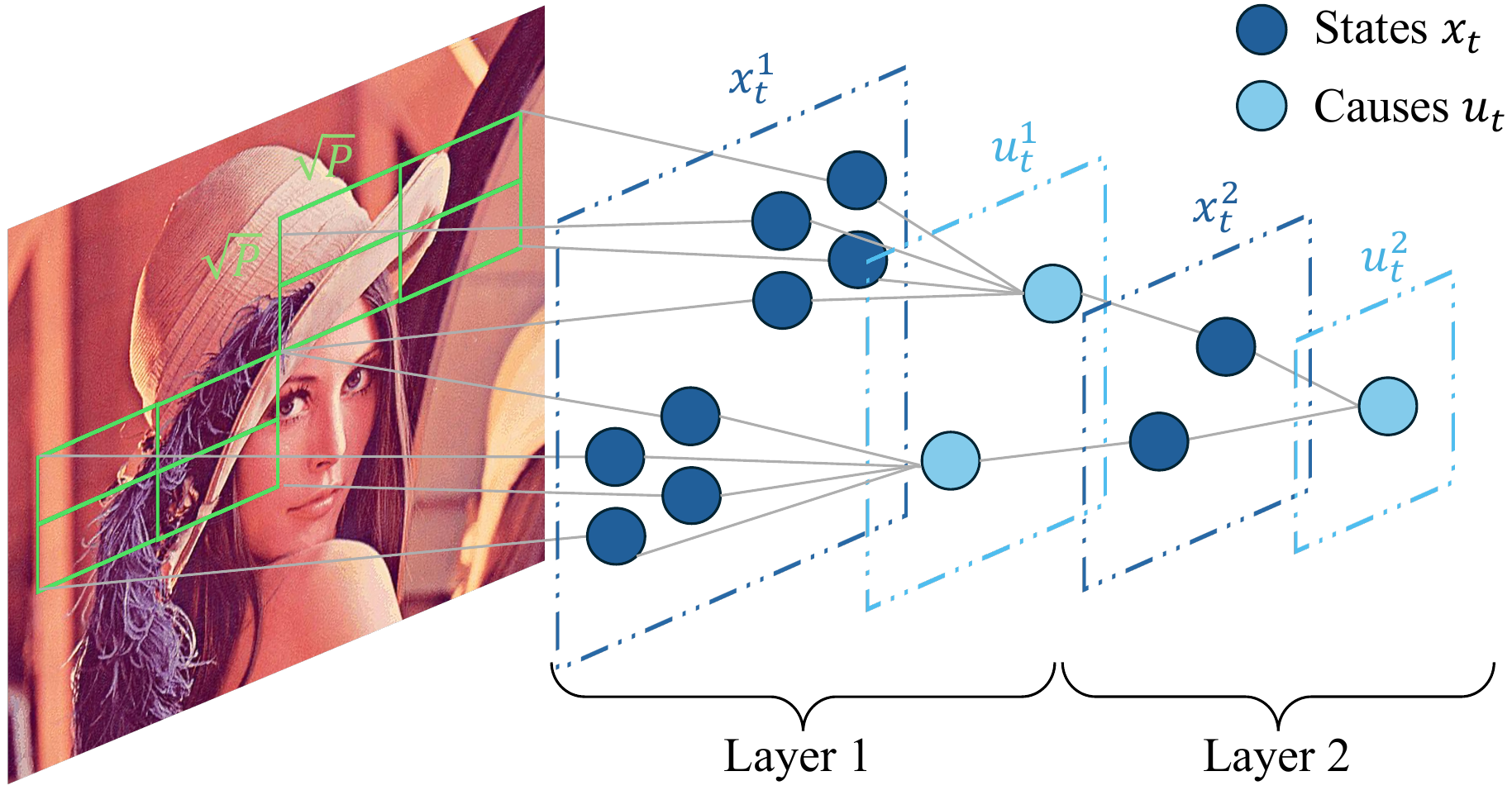}
    \caption{Two-layered DPCNs structure. The video frame is decomposed into patches (green blocks). Every patch is mapped onto a state $x_t^1$ at layer 1, and the cause $u_t^1$ pool all the states within a group. The cause $u_t^1$ is input of layer 2 and corresponds to state $x_t^2$ and cause $u_t^2$.
    }
    \vspace{-3em}
    \label{fig1}
\end{wrapfigure}
The structure of DPCN is shown in Fig. \ref{fig1}, and the involved denotations are show in Table \ref{tab:denotations}. Given a video input, the measurements of each video frame are decomposed into multiple contiguous patches in terms of position, which is denoted by $y_{t,n} \in \mathbb{R}^P,  n \in \mathcal{N}=\{1,2,\cdots,N\}$, a vectorized form of $\sqrt{P} \times \sqrt{P}$ square patch. These measurements are injected to the DPCNs with a hierarchical multiple-layered structure. From the second layer, the causes from a lower layer serve as the input of the next layer, i.e., $y_{t,n}^l=u_{t,n}^{l-1}$. At every layer, the network consists of two distinctive parts: feature extraction (inferring states) and pooling (inferring causes). {The parameters to connect states and causes are called model (dictionary), going along states and causes (inferring model). The networks and connections at each layer $l$ are given in terms of objective functions for the inferences. In the following, we would omit the layer superscript $l$ for simplicity. }

For inferring states given a patch measurement $y_{t,n}$, a linear state space model using an over-complete dictionary of $K$-filters, i.e., $C \in  \mathbb{R}^{P \times K}$ with $P < K$, to get sparse states $x_{t,n} \in \mathbb{R}^K$. Also, a state-transition matrix $A \in \mathbb{R}^{K\times K}$ is applied to keep track of historical sparse states dynamics. To this end, the objective function for states is given by
\begin{align}
E_x (x_t)=&\sum_{n=1}^{N}  \frac{1}{2} \|y_{t,n}-Cx_{t,n} \|_2^2+\mu \|x_{t,n}\|_1\nonumber\\
&+ \lambda \|x_{t,n}-Ax_{t-1,n}\|_1,
\label{10}
\end{align}
where $\lambda>0$ and $\mu>0$ are weighting parameters, $\| \cdot \|_2^2$ is the $L_2$-norm denoting energy, and $\|\cdot\|_1$ is the $L_1$-norm serving as the penalty term to make solution sparse \cite{l0}. 

For inferring causes given states, $u_t\in \mathbb{R}^{D}$ multiplicatively interacts with the accumulated states through $B\in \mathbb{R}^{K\times D}$ in the way that whenever a component in $u_t$ is active, the corresponding set of components in $x_t$ are also likely to be active. This is for significant clustering of features even with non-stationary distribution of states \cite{bay}. To this end, the objective function for causes is given by
\begin{align}
E_u (u_t)=&\sum_{n=1}^{N} \sum_{k=1}^{K} \gamma |(x_{t,n})_k| (1+exp(-(Bu_t)_k))+\beta \|u_t\|_1
\label{13}
\end{align}
where $\gamma>0$ and $\beta>0$ are weighting parameters. 

For inferring model $\theta=\{A, B, C\}$ given states and causes, the overall objective function is given by
\begin{align}
&E_p (x_t, u_t, \theta)=E_x (x_{t}) +E_u (u_t).
\label{12}
\end{align}
{Notably, optimization of the functions $E_x$ and $E_u$ are strong convex problems, and we will design learning method to find the unique optimal sparse solution.}

\section{Learning For Model Inference and Variable Inference}

\begin{wrapfigure}{1}[0cm]{0pt}
\vspace{-10em}
    \includegraphics[width=0.48\textwidth]{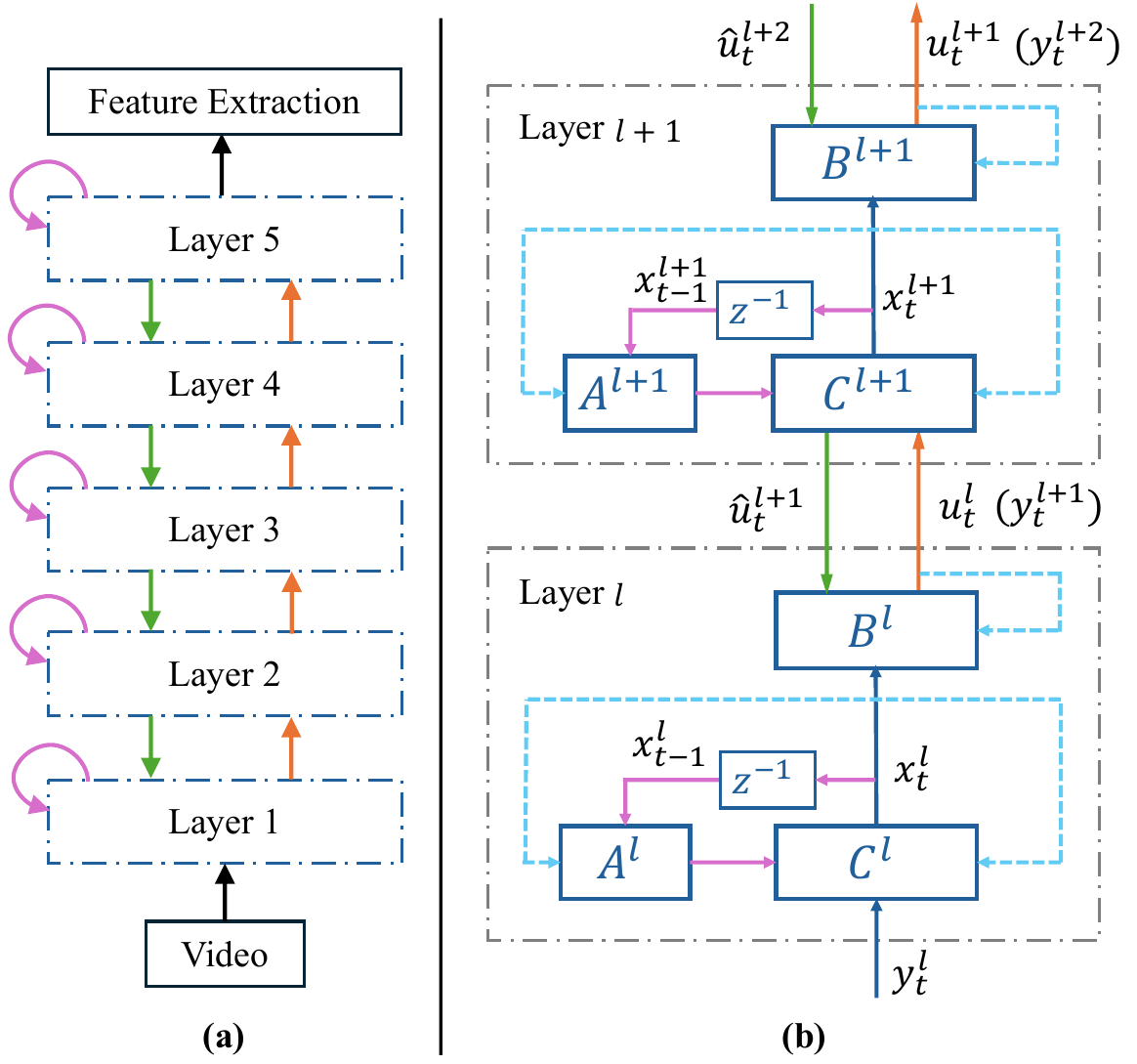}
    \caption{(a) Bi-directional inference flow, where feedforward (yellow), feedback (green), and recurrent (pink) connections convey the bottom-up and top-down predictions. (b) Connections for variables inference (solid lines) and for model inference (dash lines).}
    \vspace{-2.4em}
\label{fig2}
\end{wrapfigure}
In this section, we propose an unsupervised learning method for self-organizing models and variables with accelerated learning while maintaining high sparsity and accuracy of feature extraction. The flow and connections for the inference are shown in Fig. \ref{fig2}. {The inference process includes Model Inference and Variable Inference. The model inference needs repeated interleaved updates on states and causes and updates on model. Then, given a model, the variable inference needs an interleaved updates on states and causes using an extra top-down preference from the upper layer. These form a bi-directional inference process on a bottom-up feedforward path, a top-down feedback path, and a recurrent path.}

For the updates of states and causes involved in the Model Inference and Variable Inference, we propose a new learning procedure using the majorization minimization (MM) framework \cite{shrinkage,mm1} for optimization with sparsity constraint. Different from the frequently used proximal gradient descent methods {iterative shrinkage-thresholding algorithm (ISTA)} and fast ISTA (FISTA) \cite{FISTA,ISTA1,ISTA2} that use a majorizer for the differentiable non-sparsity-penalty terms \cite{dpcn}, this paper uses a majorizer for sparsity penalty. As such the convex non-differentiable optimization problem with sparsity constraint is transformed into a convex and differentiable problem. Moreover, taking advantage of over-complete dictionary and the iteration form inspired by the value iteration of RL \cite{adp}, the iterations for inference are derived from the condition for the optimal sparse solution to MM-based optimization problems. This also differs from the traditional gradient descent method and adaptive moment estimation (ADAM) \cite{adam} method for solving optimization problems.

\subsection{MM-Based Model Inference}

Model inference seeks $\theta=\{A, B, C\}$ by minimizing $E_p (x_t, u_t, \theta)$ in \eqref{12} with an interleaved procedure to infer states and causes by minimizing $E_x$ \eqref{10} and $E_u$ \eqref{13}.

\paragraph{State Inference}

To infer sparse $x_{t,n}$ by minimizing $E_x$ \eqref{10}, first, we let $e_{t,n}=x_{t,n}-Ax_{t-1,n}$ and use the Nesterov’s smooth approximator \cite{Smooth1,Smooth2} taking the form
\begin{align}
&\|e\|_1 \approx  f_s(e)\triangleq (\alpha^*)^T e - \frac{m}{2}\|\alpha^*\|_2^2 
\label{x2}
\end{align}
where $m>0$ is a constant and $\alpha^*$ is some vector reaching the best approximation. Then, we find a majorizer for the penalty term $\mu\|x_{t,n}\|_1$ \cite{shrinkage} in the form   
\begin{align}
\mu \|x\|_1\leq h(x,V_x)\triangleq \frac{1}{2}x^TW_x x+c 
\label{x7}
\end{align}
with equality at $x=V_x$, where $V_x$ is a vector, $W_x=\text{diag}(\mu./|{V_x}|)$ with $./$ a component-wise division product, and $c$ is a constant independent of $x$ (see details in Appendix \ref{A1}). 

Applying the approximator \eqref{x2}, majorizer \eqref{x7} and MM principles, the minimization problem of $E_x$ \eqref{10} can be transformed to the minimization of
\begin{align}
&H_x(x_{t,n})=\sum_{n=1}^{N}  \frac{1}{2}\| y_{t,n}-Cx_{t,n} \|_2^2+ \lambda f_s(e_{t,n}) +h(x_{t,n},V_x).
\label{x9}
\end{align}
Minimizing $H_x$ with respect to $x_{t,n}$ yields the Karush–Kuhn–Tucker (KKT) condition for the optimal sparse state
\begin{align}
(C^TC+ W_x)x_{t,n}= C^Ty_{t,n}-\lambda \alpha^*.
\label{x10}
\end{align}
To find such an optimal state, we propose Algorithm \ref{al2} that is applicable for every layer, applying an iterative form of \eqref{x10}. The update of states at each iteration is one-step optimal. 
We set a positive initial value for state. Note that it cannot be zero because the iteration will never update with $R_x^0=0$. Also, the optimal state in \eqref{x10} is expected to be sparse, namely some components of $x_{t,n}^i$ go to zero. This makes entries of $W_{x}$ go to infinity, leading to numerically inaccurate results. We avoid this by using $R_x=(W_x)^{-1}$ and the matrix inverse lemma \cite{mm2}
\begin{align}
(C^TC+W_x)^{-1} &=R_x - R_x C^T (I+CR_xC^T)^{-1} C R_x\triangleq T(C,R_x).
\label{x14}
\end{align}
Note that the matrix $C^TC + W_x$ is invertible due to positive semi-definite $C^TC$ and positive definite diagonal $W_x$. To further accelerate the computation, we can avoid directly computing the inverse term \((I + C R_x C^T)^{-1}\) by using the conjugate gradient method to compute \((I + C R_x C^T)^{-1} C R_x\).

\paragraph{Cause Inference}

To infer sparse causes by minimizing $E_u$ \eqref{13}, we find a majorizer of $\beta \|u_t\|_1$ as 
\begin{align}
&\beta \|u\|_1 \leq h(u,V_u)\triangleq \frac{1}{2}u^TW_u{u}+c 
\label{c6}
\end{align}
with equality at $u=V_u$, where $W_u=\text{diag}(\beta./|{V_u}|)$. Therefore, based on MM principles, we transform the minimization of $E_u$ in \eqref{13} to the minimization of
\begin{align}
H_u(u_t) = |X_t|^T (1+exp(-Bu_t)) + h(u_t,V_u)
\label{c9}
\end{align}
where $|X_t|=\gamma \sum_{n=1}^{N} |x_{t,n}|$. Minimizing $H_u$ with respect to $u_t$ yields the KKT condition
\begin{align}
W_u u_t=B^T(|X_t|. exp(-Bu_t)).
\label{c10}
\end{align}
To find such an optimal cause, we propose Algorithm \ref{al3} that is applicable for every layer, applying the iterative form of \eqref{c10} for causes inference. Since $R_u=(W_u)^{-1}$ and the iteration never update with $R_u^0=0$, we set an initial value $u_t^0>0$.

With fixed model parameter $\theta$, states $x_{t,n}$ and causes $u_t$ can be updated interleavely until they converge. Since sparsity penalty terms are replaced by a majorizer in the learning, small values of the variables are clamped via thresholds, $e_x>0$ for states and $e_u>0$ for causes, to be zero. As such, the states and causes become sparse at finite iterations.

\vspace{-1em}
\begin{minipage}{.48\textwidth}
\begin{algorithm}[H]
\caption{State Inference}\label{al2}
\textbf{1. Initialization:} initial values of states {$x_{t,n}^{0}$}, initial iteration step {$i= 0$}.

\textbf{2. Update State at patch $n$ and time $t$}  
\begin{align}
&x_{t,n}^{i+1}=T(C,R_x^i) (C^Ty_{t,n}-\lambda \alpha^*), \label{x15}\\
&R_x^i=\text{diag}(  \frac{|x_{t,n}^i|}{\mu}),\label{x16}
\end{align}

\textbf{3.} Set $i=i+1$ and repeat 2 until it converges.
       
\end{algorithm}
\end{minipage}
\hspace{0.4cm}
\begin{minipage}{.48\textwidth}
\begin{algorithm}[H]
\caption{Cause Inference}\label{al3}
\textbf{1. Initialization:} initial values of causes {$u_{t}^{0}$}, initial iteration step {$j= 0$}. 

\textbf{2. Update Causes at time $t$}: 
\begin{align}
&u_t^{j+1}=R_u^jB^T( |X_t|. exp(-Bu_t^{j})), \label{c11}\\
&R_u^j=\text{diag}( \frac{|u_t^j|}{\beta}).
\label{c12}
\end{align}

\textbf{3.} Set $j=j+1$ and repeat 2 until it converges.
       
\end{algorithm}
\end{minipage}
\vspace{-1em}

\paragraph{Model Parameters Inference}

By fixing the converged states and causes, the model parameters $\theta=\{A, B, C\}$ are updated based on the overall objective function \eqref{12}. For time-varying input, to keep track of parameter temporal relationships, we put an additional constraint on the parameters \cite{dpcn5,dpcn}, i.e., ${\theta_t}=\theta_{t-1}+z_t$,
where $z_t$ is Gaussian transition noise as an additional temporal smoothness prior. Along with this constraint, each matrix can be updated independently using gradient descent. It is encouraged to normalize columns of matrices $C$ and $B$ after the update to avoid any trivial solution.

\subsection{MM-Based Variable Inference with Top-Down Preference}

Given the learned model, the updates of states and causes in variable inference process are the same as Section IV-A except for adding $E_u$ \eqref{13} with a top-down preference for causes inference. Since the causes at a lower layer serves as the input of an upper layer, therefore, a predicted top-down reference using the states from the layer above is injected into causes inference of the lower layer. That is,
\begin{align}
\bar E_u (u_t)=&E_u(u_t)+\frac{1}{2}\|u_t-\hat u_t \|_2^2,
\label{13a}
\end{align}
where $\hat u_t$ is the top-down prediction \cite{fast}. Determination of its value can be found in Appendix \ref{A1} and \cite{dpcn}. Similar to Section 3.1, using the majorizer \eqref{c6} to replace the $L_1$-norm penalty in $E_u$, minimizing $\bar E_u$ \eqref{13a} becomes minimizing
\begin{align}
\bar H_u(u_t)=H_u(u_t)+\frac{1}{2}\|u_t-\hat u_t \|_2^2.
\label{19}
\end{align}
with respect to $u_t$, which yields the KKT condition
\begin{align}
(I+W_{u})u_t=\hat u_t+B^T(|X_t|. exp(-Bu_t))
\label{17a}
\end{align}
for every layer, where $I$ denotes identity matrix. Since the diagonal matrix $(I+W_{u})$ is non-singular, we develop the iterative form in Algorithm \ref{al4}.

Since inferences at each layer are independent, the complete learning procedure for each layer is summarized in Algorithm \ref{al1}. For better convergence of state inference and cause inference that are interleaved in an alternating minimization manner, we encourage to run Algorithm \ref{al2} for several iterations $i_s$ and then Algorithm \ref{al3} for several iterations $j_s$.

\begin{minipage}{.48\textwidth}
\begin{algorithm}[H]
\caption{Top-down Cause Inference}\label{al4}
\textbf{1. Initialization:} initial values of causes $u_{t}^{0}$, initial iteration step {$j= 0$}.\\
\textbf{2. Update Causes at time $t$}: 
\begin{align}
&u_t^{j+1}=T(I_D,\bar R_u^j) \left(\hat u_t+(B)^T\right.\nonumber\\
& \quad \quad \quad \ \times \left. ( |X_t|. exp(-Bu_t^{j})) \right) \label{17}\\
&\bar R_u^j=\text{diag}( \frac{|u_t^{j}|}{\beta}  ).
\label{18}
\end{align}

\textbf{3.} Set $j=j+1$ and repeat 2 until it converges.
       
\end{algorithm}

\end{minipage}
\hspace{0.4cm}
\begin{minipage}{.48\textwidth}
\begin{algorithm}[H]
\caption{MM-DPCNs}\label{al1}
\textbf{1. Initialization:} Video input {$y_{t,n}$}, initial model parameters {$\theta^{0}$}, initial variables {$x_{t,n}^{0}, u_{t}^{0}$}.  

\textbf{2. Model Inference}: \\
      i). Update state $x_{t,n}$ by Algorithm \ref{al2} and cause $u_{t}$ by Algorithm \ref{al3} interleavely until converge.\\
      ii). Update dictionary $\theta$ using gradient descent method once.\\
      iii) Go to step i) until $\theta$ converges.

 \textbf{3. Bi-Directional Variable Inference:}  \\
Fix model $\theta$. Run Algorithms \ref{al2} and \ref{al4} interleavely to infer $x_{t,n}$ and $u_{t}$ until they converge. 
\end{algorithm}
\end{minipage}

\section{Convergence Analysis of MM-Based Variable Inference}

In this section, we analyze the convergence of the proposed Algorithm \ref{al2} for state inference and Algorithm \ref{al3} for cause inference, respectively. 

\paragraph{Convergence of State Inference}

States inference is independent at each patch $n$ and each layer $l$, hence we analyze the convergence of the objective function of $E_x$ \eqref{10} using Algorithm \ref{al2} by removing the subscript $n$ and $l$ for simplicity. To do this, we introduce an auxiliary objective function
\begin{align}
&F(x_t) = f(x_t) + g(x_t) 
\label{a1}
\end{align}
where $f(x_t) = \frac{1}{2}\| y_{t}-Cx_{t} \|_2^2+ \lambda f_s(e_{t})$ and $g(x_t) = \mu \| x_{t} \|_1$.
Rewrite $H_x$ in \eqref{x9} for each patch as
\begin{align}
H_x(x_{t},V_t) = f(x_t) + h(x_t, V_x)
\label{a3}
\end{align}
where $g(x_t) \leq h(x_t, V_x)$ with equality at $x_t=V_x$ as shown in \eqref{x7}. This admits the unique minimizer 
\begin{align}
&P(V_x): = \underset{x_t}{\text{argmin}} H_x(x_{t},V_x).
\label{a6}
\end{align}

\begin{theorem} \label{th1}
Consider the sequence $\{x_t^{i}\}\in \mathbb{R}^K$ for a patch generated by Algorithm \ref{al2}. Then, $F(x_t^i)$ converges, and for any $s\geq 1$ we have
\begin{align}
F(x_t^{s})-F(x_t^*) \leq \frac{1}{2s} \sum_{i=0}^{s-1}  (|x_t^*|-|x_t^i|)^TR(|x^*|-|x_t^i|)
\label{t11}
\end{align}
where $R=\text{diag} \{1/(\tilde {{1}}|(x_t^{0})_k|+(1-\tilde {{1}}-\bar {{1}})|(x_t^{*})_k|+\bar {{1}}|(x_t^{i})_k|)\}$, 
$k\in \{1,2,..., K\}$, with $\tilde {{1}}=1$ if $|(x^*)_k| \geq |(x_t^{0})_k|>0$, $\tilde {{1}}=0$ if $0 \leq |(x^*)_k| < |(x_t^{0})_k|$, $\bar {{1}}=1$ if $|(x^*)_k|=0$, and $\bar {{1}}=0$ otherwise. Notably, $()_k$ denotes the $k$-th elements of a vector.
\end{theorem}
\textbf{Proof:} Please see Appendix \ref{A2}.

\begin{theorem} \label{t2}
Let $x_t^*$ be the optimal solution to minimizing $E_x$ \eqref{10} for a single patch at a layer. The upper bound of its convergence satisfies
\begin{align}
E_x(x_t^{s})-E_x(x_t^*) \leq \lambda m\bar D+ \frac{1}{2s} \sum_{i=0}^{s-1}  (|x_t^*|-|x_t^i|)^TR(|x_t^*|-|x_t^i|).
\label{t21}
\end{align}
where $\bar D=\underset{\|\alpha\|_\infty \leq 1}{\text{max}}  \frac{1}{2}\|\alpha\|_2^2 $.
\end{theorem}
\textbf{Proof:} Please see Appendix \ref{A2}.

\paragraph{Convergence of Causes Inference}
The convergence of cause inference can be analyzed similarly. We rewrite the function $E_u$ \eqref{13} at a single layer as
\begin{align}
E_u(u_t)=f_u(u_t) +\beta \|u_t\|_1
\label{a29}
\end{align}
where $f_u(u_t)=|X_t|^T (1+exp(-Bu_t))$. We also rewrite $H_u$ \eqref{c9} with \eqref{c6} as
\begin{align}
H_u(u_t,V_u)=f_u(u_t) + h(u_t,V_u).
\label{a30}
\end{align}

\begin{theorem} \label{t3}
Consider the sequence $\{u_t^{j}\}\in \mathbb{R}^D$ generated by Algorithm \ref{al3}. Then, $E_u(u_t^j)$ converges, and for any $s\geq 1$ we have
\begin{align}
E_u(u_t^{s})-E_u(u_t^*) \leq \frac{1}{2s} \sum_{j=0}^{s-1}  (|u_t^*|-|u_t^j|)^T\bar R(|u_t^*|-|u_t^j|).
\label{a32}
\end{align}
where $\bar R=\text{diag} \{1/(\tilde {1}|(u_t^{0})_k|+(1-\tilde{1}-\bar {{1}})|(u_t^{*})_k|+\bar {{1}}|(u_t^{j})_k|)\}$, $k\in \{1,2,..., D\}$, with $\tilde{1}=1$ if $|(u_t^*)_k| \geq |(u_t^{0})_k|>0$, $\tilde{1}=0$ if $0\leq |(u_t^*)_k| < |(u_t^{0})_k|$, $\bar {{1}}=1$ if $|(u_t^*)_k|=0$, and $\bar {{1}}=0$ otherwise.
\end{theorem}
\textbf{Proof:} Please see Appendix \ref{A2}.

We have a similar conclusion for Algorithm \ref{al4}. In Algorithm \ref{al4}, we set initial $u_t^0> 0$. With a diagonal positive-definite matrix $T(I_D, \bar R_u^j)$, i.e., $(I+W_{u}^j)^{-1}$, given $u_t^{j}>0$, \eqref{17} with a normalized matrix $B$ yields $u_t^{j+1}>0$. Using similar proof of Algorithm \ref{al3}, we can induce that Algorithm \ref{al4} will make $u_t^j$ sparse and minimizes $\bar H_u$ in \eqref{19}. Based on the MM principles, it also minimizes the function $\bar E_u$ in \eqref{13a}.

\section{Experiments}

We report the performance of MM-DPCNs on image sparse coding and video feature clustering. We compare MM-based algorithm used for MM-DPCNs with the methods FISTA \cite{FISTA}, ISTA \cite{ISTA1}, ADAM \cite{adam} to test optimization quality of sparse coding on the CIFAR-10 data set. For video feature clustering, we compare our MM-DPCNs to previous DPCNs version FISTA-DPCN \cite{dpcn} and methods auto-encoder (AE) \cite{AE}, WTA-RNN-AE \cite{WTARNN} (architecture details are provided in Appendix \ref{res-AE-arch}) on video data sets OpenAI Gym Super Mario Bros environment \cite{openai_super_mario_bros} and Coil-100 \cite{coil100}. Note that these are the standard data sets used for sparse coding and feature extraction \cite{hongming,Qian}. We use indices including clustering accuracy (ACC) as the completeness score, adjusted rand index (ARI) and the sparsity level (SPA) to evaluate the clustering quality, learning convergence time (LCT) for sparse coding optimization on each frame. More results on a geometric moving shape data set can be found in Appendix \ref{res-AE-arch}. The implementations are written in PyTorch-Python, and all the experiments were run on a Linux server with a 32G NVIDIA V100 Tensor Core GPU.

\begin{figure*}[htb]
    \centering
    \begin{subfigure}{0.32\textwidth}
        \centering
        \includegraphics[width=0.99\linewidth]{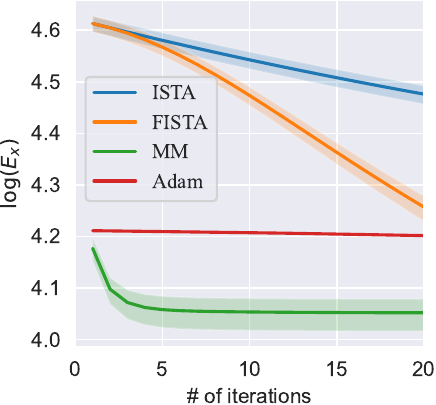}
        \caption{Convergence.}
        \label{convergence}
    \end{subfigure}
    \begin{subfigure}{0.32\textwidth}
        \centering
        \includegraphics[width=\linewidth]{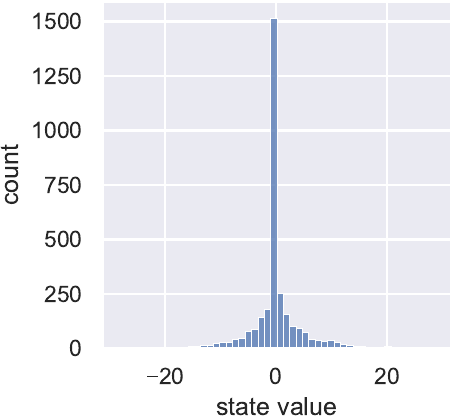}
        \caption{MM state histogram.}
        \label{MM-State}
    \end{subfigure}
    \begin{subfigure}{0.32\textwidth}
        \centering
        \includegraphics[width=0.97\linewidth]{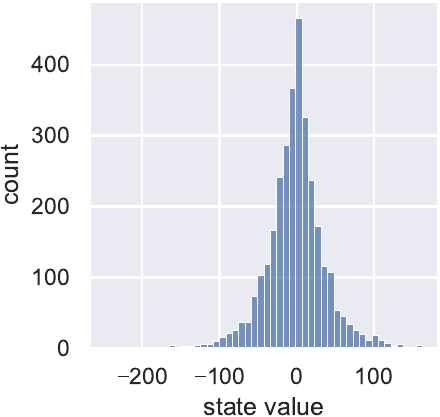}
        \caption{FISTA state histogram.}
        \label{FISTA-State}
    \end{subfigure}
    \caption{(a) Convergence of MM Algorithm \ref{al2}, ISTA, FISTA, and ADAM, (b) sparsity level using MM Algorithm \ref{al2}, and (c) sparsity level using FISTA.}
    \vspace{-1em}
    \label{iteration}
\end{figure*}

\subsection{Comparison on Image Sparse Coding}

\begin{wraptable}{r}{6.5cm}
\centering
    \caption{CIFAR-10 sparse coding optimization.}
    \begin{tabular}{cccccccccccccc}
    \toprule\midrule
    \multirow{1}{*}{Methods} & \multicolumn{1}{c}{$E_x$} & \multicolumn{1}{c}{SPA}\\
    \midrule\midrule
    ISTA & $\expnum{2.96}{4}\pm680$ & $8.96\pm0.39$\\
    FISTA & $\expnum{1.77}{4} \pm 537$ & $19.50\pm0.83$\\
    Adam & $\expnum{1.59}{4}\pm 13.68$ & $34.99\pm0.05$\\
    MM & $\expnum{1.09}{4}\pm390$ & $79.87\pm0.32$ \\
    \midrule\bottomrule
    \end{tabular}
\vspace{-1.5em}
\label{table1}
\end{wraptable}
The proposed MM Algorithms \ref{al2} is applicable for general sparse optimization problems such as Lasso problems \cite{lasso}. We apply the MM Algorithm \ref{al2}, as well as the well-known ISTA \cite{ISTA1}, FISTA \cite{FISTA} for comparison, on the CIFAR-10 data set with the reconstruction and sparsity loss $E_x$ \eqref{10} ($\mu=0.3$, $\lambda=0$, and randomized $C\in\mathbb{R}^{256\times300}$). {We also compare the performance with the Adam algorithm \cite{adam} to optimize the smooth majorizer, which is of particular interest to the Deep Learning optimization community.}  The images are preprocessed by splitting into four equally-sized patches. FISTA and ISTA have learning rates, set as $\eta=1e-2$, while MM is learning-rate-free. 

Fig.~\ref{convergence} shows that the MM Algorithm \ref{al2} converges in less than 10 steps, much faster than the others. Also, it enjoys a higher sparsity level of the learned state, {to be a direct benefit of fast convergence rate}, as shown in Fig. \ref{MM-State} and Fig. \ref{FISTA-State}. 
The statistics of the optimization results are summarized in Table \ref{table1}, where MM Algorithm \ref{al2} produces the least loss value while maintaining the highest sparsity level. The results reveal three potential advantages for MM-DPCN: 1. Faster computation. 2. Higher level sparsity for the latent space embeddings. 3. More faithful reconstructions. The last two advantages enable the algorithm to produce highly condensed and faithful information embedded into the latent space, which also benefits feature clustering.

\subsection{Comparison on Video Clustering}

\paragraph{Super Mario Bors data set}

\begin{figure*}[htb]
    \centering
    \begin{subfigure}{0.24\textwidth}
        \centering
        \includegraphics[width=\linewidth]{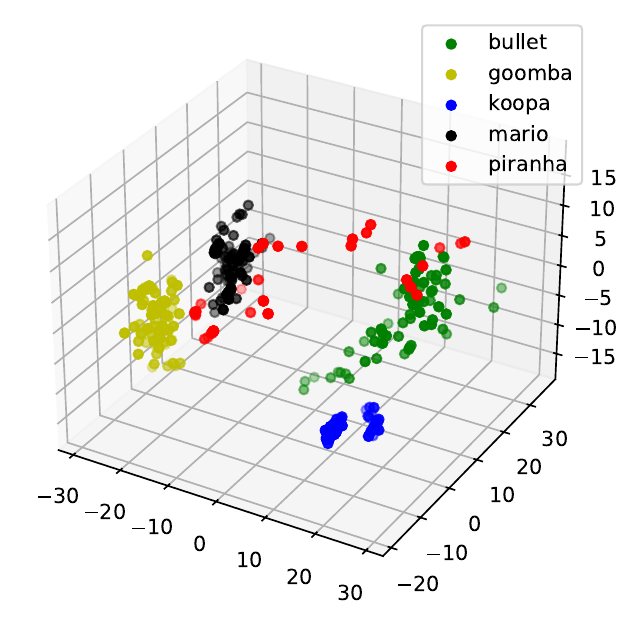}
        \caption{AE}
        \label{AE}
    \end{subfigure}
    \begin{subfigure}{0.24\textwidth}
        \centering
        \includegraphics[width=\linewidth]{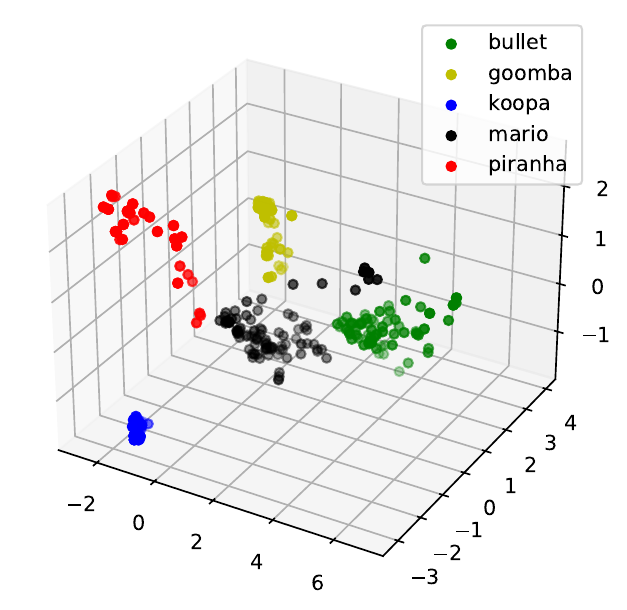}
        \caption{WTA-RNN-AE}
        \label{WTA}
    \end{subfigure}
    \begin{subfigure}{0.24\textwidth}
        \centering
        \includegraphics[width=\linewidth]{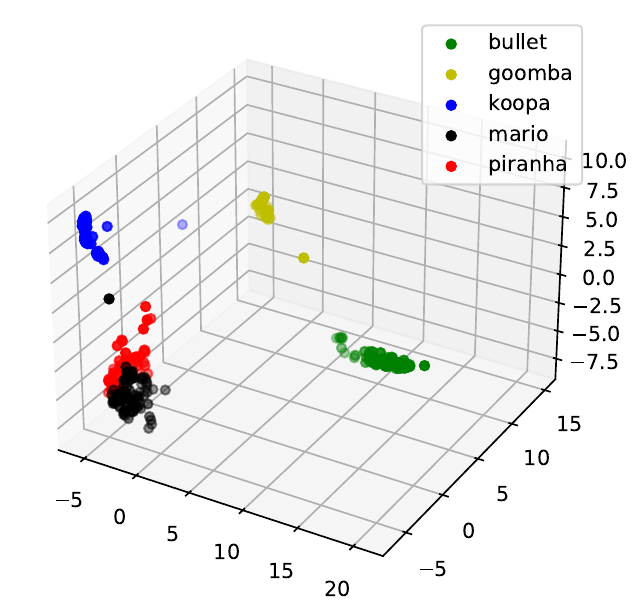}
        \caption{FISTA-DPCN}
        \label{FISTA}
    \end{subfigure}
    \begin{subfigure}{0.24\textwidth}
        \centering
        \includegraphics[width=\linewidth]{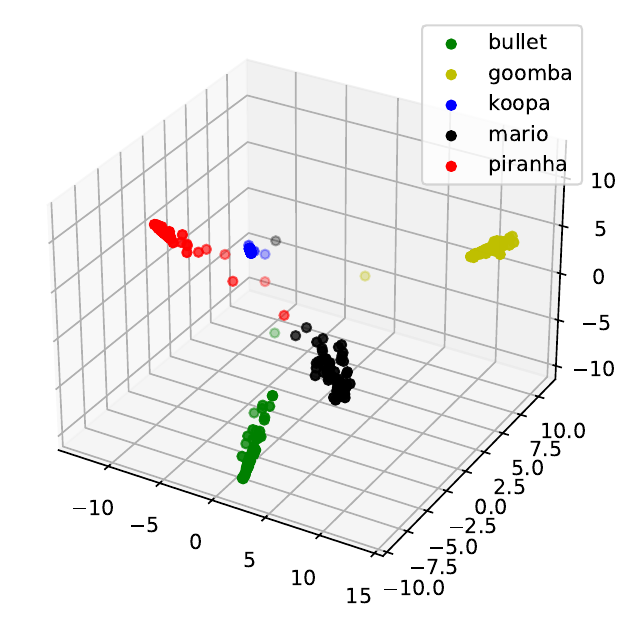}
        \caption{MM-DPCN}
        \label{MM}
    \end{subfigure}
    \caption{Clustering result for a Super Mario Bros video data set. }
    \label{clusters}
\end{figure*}

\begin{figure}
    \centering
    \begin{subfigure}{\textwidth}
        \raisebox{-\height}{\includegraphics[width=\linewidth]{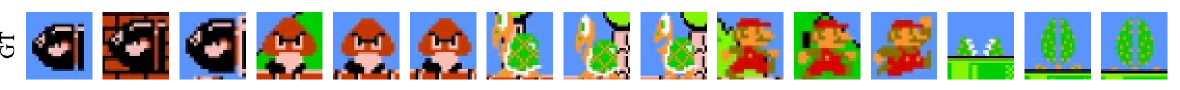}}\vspace{-0.5em}
        \raisebox{-\height}{\includegraphics[width=\linewidth]{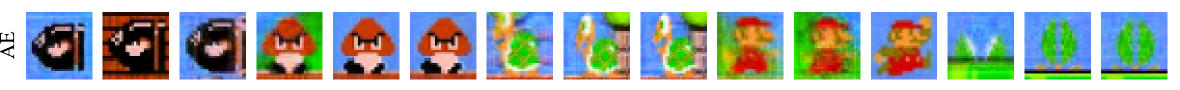}}\vspace{-0.5em}
        \raisebox{-\height}{\includegraphics[width=\linewidth]{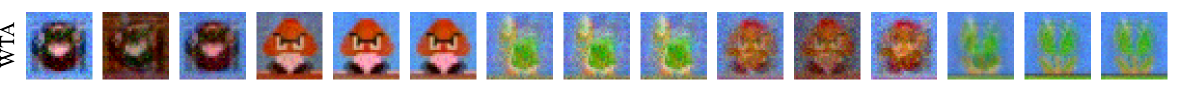}}\vspace{-0.5em}
        \raisebox{-\height}{\includegraphics[width=\linewidth]{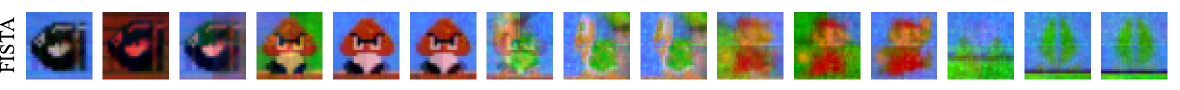}}\vspace{-0.5em}
        \raisebox{-\height}{\includegraphics[width=\linewidth]{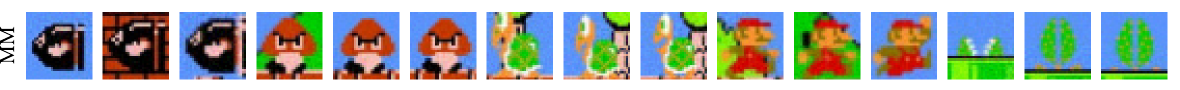}}\vspace{-0.5em}
        \caption{Super Mario Bros}
        \label{mario-recon}
    \end{subfigure}
    \begin{subfigure}{\textwidth}
        \raisebox{-\height}{\includegraphics[width=\linewidth]{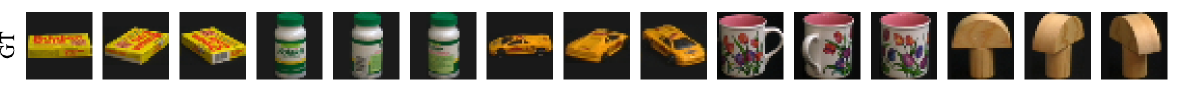}}\vspace{-0.5em}
        \raisebox{-\height}{\includegraphics[width=\linewidth]{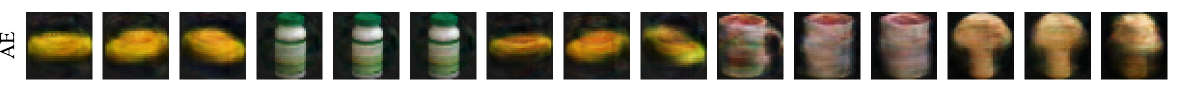}}\vspace{-0.5em}
        \raisebox{-\height}{\includegraphics[width=\linewidth]{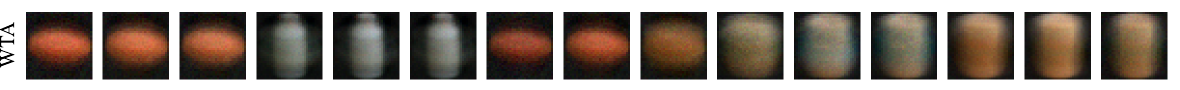}}\vspace{-0.5em}
        \raisebox{-\height}{\includegraphics[width=\linewidth]{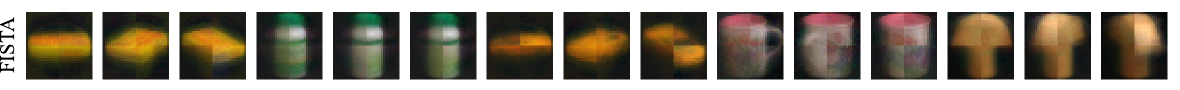}}\vspace{-0.5em}
        \raisebox{-\height}{\includegraphics[width=\linewidth]{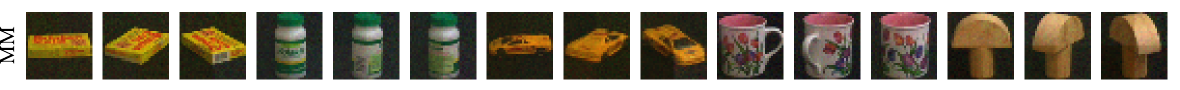}}
        \caption{Coil-100}
        \label{coil-100-recon}
    \end{subfigure}
    \caption{Qualitative video sequence reconstruction for Super Mario Bros and Coil-100 data sets. }
    \vspace{-1em}
    \label{recons}
\end{figure}

We picked five main objects of the Mario \cite{openai_super_mario_bros} data set from the video sequence played by humans: Bullet Bill, Goomba, Koopa, Mario, and Piranha Plant. They exhibit various movements, such as jumping, running, and opening or closing, against diverse backgrounds. Both training and testing videos contain 500 frames ($32\times32\times3$ pixels), with 100 consecutive frames per object. For DPCNs, each frame is divided into four vectorized patches normalized between 0 and 1. It is initialized with  $x^1\in\mathbb{R}^{300}$, $u^1\in\mathbb{R}^{40}$, $x^2\in\mathbb{R}^{100}$, $u^2\in\mathbb{R}^{20}$, and model matrices $A^l, B^l, C^l$, $l=1,2$. We set $\mu^l=0.3$ and $\beta^l=0.3$ for MM-DPCN and $\mu^l=1$ and $\beta^l=0.5$ for FISTA-DPCN.  
Figure \ref{clusters} shows that MM-DPCN produces a clean separation while keeping each cluster compact. Figure \ref{mario-recon} demonstrates the optimal reconstruction quality produced by MM-DPCN in comparison to alternative methods. We obseve from {Table \ref{tab:clustering-comparison} that MM-DPCN achieves the best ACC, ARI, SPA, and is much faster than previous version FISTA-DPCN.}

\paragraph{Coil-100 data set}

The Coil-100 data set \cite{coil100} consists of 100 videos of different objects, with each 72 frames long. The frames are resized into 32×32 pixels and normalized between 0 and 1. We used the first 50 frames of all the objects for training, while the rest 22 frames for testing. We initialize our MM-DPCNs with randomized model $A^l, B^l, C^l$, $l=1,2$, and $x^1\in\mathbb{R}^{2000}$, $x^2\in\mathbb{R}^{500}$, $u^1\in\mathbb{R}^{128}$ and $u^2\in\mathbb{R}^{80}$. We set $\mu^l=0.1$, $\beta^l=0.1$ for MM-DPCN and $\mu^l=1$, $\beta^l=0.2$ for FISTA-DPCN. We extract the causes from the last layer of MM- and FISTA-DPCNs and use PCA to project them into three-dimensional vectors, then apply K-Means for clustering. This same process is applied to the learned latent space encodings for both AE and WTA-RNN-AE, constructed using MLPs and ReLU. 

\begin{table}[ht]
\vspace{-1em}
    \centering
    \caption{Quantitative comparison for video clustering and learning convergence time.}
    \begin{tabular}{ccccccccc}
    \toprule\midrule
    \multirow{2}{*}{Methods} & \multicolumn{4}{c}{Mario} & \multicolumn{4}{c}{Coil-100}\\
    \cmidrule(r){2-9}
    & ACC & ARI & SPA & LCT ($s$) & ACC & ARI & SPA & LCT ($s$) \\
    \midrule\midrule
    AE & 84.81 & 76.74 & 0.00 & * & 77.74 & 44.04 & 0.00 & *\\
    WTA-RNN-AE & 92.76 & 88.22 & 90.00 & * & 79.28 & 44.45 & \textbf{90.00} & *\\
    FISTA-DPCN & 87.74 & 72.01 & 87.22 & 0.084 & 80.48 &  47.00 & 81.02 & 0.102 \\
    MM-DPCN & \textbf{94.87} & \textbf{91.98} & \textbf{95.17} & \textbf{0.015} & \textbf{82.98} & \textbf{48.93} & 57.86 & \textbf{0.016} \\
    
    \midrule\bottomrule
    \end{tabular}
    \vspace{-1em}
    \label{tab:clustering-comparison}
\end{table}

Table \ref{tab:clustering-comparison} presents the quantitative clustering and learning results, and Figure \ref{coil-100-recon} showcases the qualitative video sequence reconstruction results. WTA-RNN-AE includes an additional RNN to learn video dynamics, which, however,  is a trade-off with reconstruction. On the other hand, the FISTA- and MM-DPCNs provide much better reconstruction as the recurrent models $A$ are linear and less susceptible to overfitting than RNN, while WTA-RNN-AE tends to blend and blur different objects. Therefore, the efficiency of the iterative process enables MM to provide the best reconstruction quality. As shown in Table \ref{tab:clustering-comparison}, WTA-RNN-AE has best SPA since it allows selected sparse level as $90\%$ for encodings, which, however, results in worse ACC and ARI due to over-loss of information. In contrast, MM and FISTA, by selecting sparsity coefficients or how much information can be compressed without resorting to nonlinear DL models, have much better ACC and ARI, where our MM-DPCN has the best ACC and ARI and MSE. 

In the learning, the matrix inversion operation involves a conjugate gradient computation with complexity approximately \( O(\sqrt{m} K^2) \), where \( m \) is the matrix condition number and \( K \) is the state size. The memory complexity for storing matrices is \( O(K^2) \), and this requirement arises as state size increases, potentially leading to memory overhead when vector size is too large. This can be mitigated to moderately increasing patches or enlarging hardware memory.

\section{Conclusion}

{We proposed a MM-based DPCNs that circumvents the non-smooth optimization problem with sparsity penalty for sparse coding by turning it into a smooth minimization problem using majorizer for sparsity penalty. The method searches for the optimal solution directly by the direction of the stationary point of the smoothed objective function. The experiments on image and video data sets demonstrated that this tremendously speeds up the rate of convergence, computation time, and feature clustering performance.}

\begin{ack}

This work is partially supported by {the Office of the Under Secretary of Defense for Research and Engineering under awards N00014-21-1-2295 and N00014-21-1-2345}
 
\end{ack}


\small

\bibliographystyle{unsrt}
\bibliography{dpcn}


\appendix

\section{Appendix for Derivations}\label{A1}

For the term $\|e_{t,n}\|_1$ where $e_{t,n}=x_{t,n}-Ax_{t-1,n}$,
the smooth approximation on it is given by
\begin{align}
&\|e_{t,n}\|_1 \approx  f_s(e_{t,n})= \underset{\|\alpha\|_\infty \leq 1}{\text{max}} \left( \alpha^T e_{t,n} - \frac{m}{2}\|\alpha\|_2^2 \right). 
\end{align}
The best approximation, as well as the maximum, is reached at $\alpha^*$ such that
\begin{align}
\alpha^*=S(\frac{e_{t,n}}{m})=\left\{
\begin{array}{cc}
    \frac{e_{t,n}}{m} & -1\leq \frac{e_{t,n}}{m}\leq 1 \\
    1 & \frac{e_{t,n}}{m}> 1 \\
    -1 & \frac{e_{t,n}}{m}< 1 
\end{array}
\right. 
\end{align}

The majorizer of the sparsity penalty is given by
\begin{align}
\mu \|x_{t,n}\|_1 \leq h(x_{t,n},V_x)= \sum_{k=1}^{K} h((x_{t,n})_k,(V_{x})_k)
\end{align}
where 
\begin{align}
h((x_{t,n})_k,(V_x)_k)
&=\frac{ \phi'((V_x)_k)}{2(V_x)_k} (x_{t,n})_k^2+\phi((V_x)_k)-\frac{(V_x)_k}{2}  \phi'(V_x)_k),\nonumber\\
&\geq \mu|(x_{t,n})_k|, \quad \  \forall (x_{t,n})_k\in \mathbb{R}. 
\end{align}
where $\phi((V_x)_k)=\mu |(V_x)_k|$ and $V_x \in \mathbb{R}^K$ can be any vector. The equality holds only at $V_x=x_{t,n}$. By rewriting the left-hand-side majorizer compactly, it becomes \eqref{x7} where $c=\sum_{k=1}^{K} \phi((V_{x})_k)-0.5{(V_x)_k}  \phi'((V_{x})_k)$ is a constant independent of $x_{t,n}$. Accordingly, the constant $c$ in \eqref{c6} is $c=\sum_{k=1}^{D} \psi((V_u)_k)-0.5{(V_u)_k} \psi'((V_u)_k)$, $\psi((u_t)_k)=\beta |(u_t)_k|$, where $V_u \in \mathbb{R}^D$ can be any vector.

{The top-down prediction for layer $l$ from the upper layer $l+1$ is denoted by $\hat u_t$ which is given by
\begin{align}
&\hat u_t^{l} =C^{l+1}\hat x^{l+1}_t,
\label{14}\\
&(\hat x^{l+1}_t)_k=\left\{
\begin{array}{cc}
    (A^{l+1}x^{l+1}_{t-1})_k & \lambda> \gamma(1+exp(-(B^{l+1} u^{l+1}_t )_k)
  \\
    0 & \lambda \leq \gamma(1+exp(-(B^{l+1} u^{l+1}_t )_k) 
\end{array}
\right.
\label{15}
\end{align}
where $\lambda$ belongs to layer $l+1$. At the top layer $L$, we set $\hat u^L_t = u^L_{t-1}$, which induces some temporal coherence on the final outputs.}

\section{Appendix for Proofs}\label{A2}

We first show a necessary lemma before proving Theorem \ref{th1}. Since $V_x$ in \eqref{x7} represents any vector with the same dimension as $x_t$, for simplification we use $V$ as $V_x$ in the following analysis regrading state inference. We also do the same, using $V$ as $V_u$ that appears in \eqref{c6}, in the analysis regrading cause inference.

\begin{lemma}\label{lm1}
    Let $V \in \mathbb{R}^K$ satisfy
    \begin{align}
    &F(P(V)) \leq H_x(P(V), V).
    \label{a10}
    \end{align}
    For any $x_t \in \mathbb{R}^K$ one has
    \begin{align}
    &F(x_t)-F(P(V)) \geq \sum_{k=1}^K -\frac{(|(x_{t})_k|-|(V)_k|)^2 }{2|(V)_k|}.
    \label{a11}
    \end{align}
\end{lemma}
\textbf{Proof:} Recalling the majorizer for states, i.e., $h(x_{t},V)$ in \eqref{x7}, it can be induced from \eqref{a3}-\eqref{a6} that $P(V)$ satisfies
\begin{align}
&\nabla f(P(V)) + \nabla_{x_t} h(P(V),V)=0.
\label{a7}
\end{align}
Then, we know from \eqref{x15} that 
\begin{align}
&x_t^{i+1} = P(x_t^i).
\label{a8}
\end{align}
It follows from \eqref{x7} that \eqref{a10} holds. Since $f(x_t)$ and $h(x_t,V_x)$ are convex on $x_t$, we have 
\begin{align}
&f(x_t) -f(P(V)) \geq \langle x_t-P(V), \nabla f(P(V_x))  \rangle, \label{a12a} \\
&h(x_t,V) - h(P(V),V) \geq \langle x_t-P(V), \nabla_{x_t} h(P(V),V) \rangle.
\label{a12b}
\end{align}
Hence, with \eqref{a10}, \eqref{a1} and \eqref{a3}, we have
    \begin{align}
    &F(x_t)-F(P(V)) \nonumber\\
    &\geq F(x_t)-H_x(P(V), V) \nonumber\\
    &=f(x_t)+g(x_t)-f(P(V))-h(P(V),V)\nonumber\\
    &\geq \langle x_t-P(V), \nabla f(P(V))  \rangle + h(x_t,x_t) - h(P(V),V)\nonumber\\
    &=\langle x_t-P(V), \nabla f(P(V))  \rangle + h(x_t,V) - h(P(V),V)  + h(x_t,x_t) - h(x_t,V) \nonumber\\
    &\geq \langle x_t-P(V), \nabla f(P(V))  \rangle + \langle x_t-P(V), \nabla_{x_t} h(P(V),V)  \rangle + h(x_t,x_t) - h(x_t,V) \nonumber\\
    &=h(x_t,x_t) - h(x_t,V).
    \label{a16}
    \end{align}
    Note that the fourth line applies \eqref{a12a} and $g(x_t)=h(x_t,x_t)$, the seventh line applies \eqref{a12b}, and the last line applies \eqref{a7}. 

    It follows from \eqref{x7} and Appendix \ref{A1} that
    \begin{align}
    h(x_{t},x_{t})-h(x_{t},V)
    &=\sum_{k=1}^K 
    (x_{t})_k \text{sign}((x_t)_k) -\frac{ \text{sign}((V)_k)}{2(V)_k} \left((x_{t})_k^2+(V)_k^2\right)\nonumber\\
    &=\sum_{k=1}^K |(x_{t})_k|-\frac{(x_{t})_k^2+(V)_k^2 }{2|(V)_k|} \nonumber\\
    &=\sum_{k=1}^K -\frac{(|(x_{t})_k|-|(V)_k|)^2 }{2|(V)_k|}\leq 0.
    \label{a14}
    \end{align}    
Substituting it into \eqref{a16} yields \eqref{a11}. This completes the proof.

\paragraph{Proof of Theorem \ref{th1}}
It can be inferred from the derivations that 
\begin{align}
    &F(x_t^i) = H_x(x_t^i,x_t^{i}) \leq  H_x(x_t^i,x_t^{i-1}) \leq  H(x_t^{i-1},x_t^{i-1}) =F(x_t^{i-1})
    \label{t12}
\end{align}
where the second and third equality hold only at $x_t^i=x_t^{i-1}$, i.e., $x_t^i$ satisfies the optimality condition \eqref{x10}. That is, $F(x_t^i)$ monotonically decreases until $x_t^i$ satisfies the optimality condition. Moreover, it follows from the approximation shown in \eqref{x2} that the approximation gap is
\begin{align}
\|e_{t,n}\|_1 - m\bar D \leq f_s(e_{t,n}) \leq \|e_{t,n}\|_1
\label{t13}
\end{align}
where $\bar D=\underset{\|\alpha\|_\infty \leq 1}{\text{max}}  \frac{1}{2}\|\alpha\|_2^2 $. It indicates that $F(x_t)$ is lower-bounded such that
\begin{align}
E_x(x_t) - \lambda m \bar D \leq F(x_t) \leq E_x(x_t) 
\label{t14}
\end{align}
where $E_x(x_t)\geq 0$. Therefore, $F(x_t^i)$ is monotonically convergent with boundaries using $E_x(x_t^i)$.

By taking $x_t=x_t^*$, $P(V)=x_t^{i+1}$, and $V=x_t^i$ in Lemma \ref{lm1}, we can write
\begin{align}
    &F(x_t^*)-F(x_t^{i+1}) \geq \sum_{k=1}^K -\frac{(|(x^*)_k|-|(x_t^{i})_k|)^2 }{2|(x_t^{i})_k|}.
    \label{t15}
\end{align}
Summing it for $s$ iterations yields
\begin{align}
    &sF(x_t^*)-\sum_{i=1}^{s} F(x_t^{i}) \geq  \sum_{i=0}^{s-1} \sum_{k=1}^K -\frac{(|(x^*)_k|-|(x_t^{i})_k|)^2 }{2|(x_t^{i})_k|}.
    \label{t16}
\end{align}
Subtracting $sF(x_t^s)$ from the both sides yields
\begin{align}
    &sF(x_t^*)-sF(x_t^s) \geq  \sum_{i=0}^{s-1} \sum_{k=1}^K -\frac{(|(x^*)_k|-|(x_t^{i})_k|)^2 }{2|(x_t^{i})_k|}+ \sum_{i=1}^{s} \left( F(x_t^{i})-F(x_t^s)\right).
    \label{t17}
\end{align}

From \eqref{t12} we infer that $\sum_{i=1}^{s} \left( F(x_t^{i})-F(x_t^s)\right) \geq 0$. Therefore, \eqref{t17} becomes 
\begin{align}
    F(x_t^s)-F(x_t^*)&\leq  \frac{1}{2s}\sum_{i=0}^{s-1} \sum_{k=1}^K \frac{(|(x^*)_k|-|(x_t^{i})_k|)^2 }{|(x_t^{i})_k|}.
    \label{t18}
\end{align}
Let $x_t^*$ be the optimal sparse solution satisfying \eqref{x10}. Since $F(x_t^i)$ is monotonically decreasing to $F(x_t^*)$, as well as the sequence $R_x^i$ in \eqref{x16}, then $|x_t^i|$ is approaching $|x_t^*|$ monotonically. Positive or negative initial $x_t^0$ does not influence result as $|x_t^0|$ is used, and the update views $x_t^0$ as positive and drives it to a non-negative $x_t^*$ and similarly, views $x_t^0$ as negative and drives it to a non-positive $x_t^*$. Note that we never choose $x_t^0=0$. Therefore, for an optimal value $(x_t^*)_k= 0$, one has 
\begin{align}
   \frac{(|(x_t^*)_k|-|(x_t^{i})_k|)^2 }{|(x_t^{i})_k|}\leq    |(x_t^{i})_k|. 
    \label{t19}
\end{align}
For an optimal value $0 < |(x_t^{0})_k| \leq |(x_t^*)_k|$, one has 
\begin{align}
   \frac{(|(x_t^*)_k|-|(x_t^{i})_k|)^2 }{|(x_t^{i})_k|}  \leq  \frac{(|(x_t^*)_k|-|(x_t^{i})_k|)^2}{|(x_t^{0})_k|}. 
    \label{t110}
\end{align}
For an optimal value $0<|(x_t^*)_k| < |(x_t^{0})_k|$, one has
\begin{align}
   \frac{(|(x_t^*)_k|-|(x_t^{i})_k|)^2}{|(x_t^{i})_k|}  \leq  \frac{(|(x_t^*)_k|-|(x_t^{i})_k|)^2}{|(x_t^{*})_k|}. 
    \label{t111}
\end{align}
Using \eqref{t19}-\eqref{t111} in \eqref{t18} for $\forall x_t^* \in \mathbb{R}^K$, we write
\begin{align}
    F(x_t^s)-F(x_t^*)
    &\leq \frac{1}{2s} \sum_{i=0}^{s-1} \sum_{k=1}^K \frac{(|(x_t^*)_k|-|(x_t^{i})_k|)^2 }{\tilde{1}|(x_t^{0})_k|+(1-\tilde{1}-\bar {\mathbbm{1}})|(x_t^{*})_k|+\bar {\mathbbm{1}}|(x_t^{i})_k|}\nonumber\\
    &=\frac{1}{2s} \sum_{i=0}^{s-1}  (|x_t^*|-|x_t^i|)^TR(|x^*_t|-|x_t^i|)
    \label{t112}
\end{align}
where $R=\text{diag} \{1/(\tilde {{1}}|(x_t^{0})_k|+(1-\tilde {{1}}-\bar {{1}})|(x_t^{*})_k|+\bar {{1}}|(x_t^{i})_k|)\}$, 
$k\in \{1,2,..., K\}$, with $\tilde {{1}}=1$ if $|(x^*)_k| \geq |(x_t^{0})_k|>0$, $\tilde {{1}}=0$ if $0\leq |(x^*)_k| < |(x_t^{0})_k|$, $\bar {{1}}=1$ if $|(x^*)_k|=0$, and $\bar {{1}}=0$ otherwise. It can be inferred from uniqueness of $x_t^*$ and monotonic convergence of $F(x_t^i)$ that the upper bound at \eqref{t112} decreases with iterations $s$. This completes the proof.

\paragraph{Proof of Theorem \ref{t2}}
We write $E_x(x_t^{s})-E_x(x_t^*)$ in three pairs as
\begin{align}
&E_x(x_t^{s})-E_x(x_t^*) = E_x(x_t^{s})-F(x_t^s)+F(x_t^s)-F(x_t^*)+F(x_t^*)-E_x(x_t^*).
\label{t22}
\end{align}
The first and third pairs in \eqref{t22}, i.e., $E_x(x_t^{s})-F(x_t^s)$ and $F(x_t^*)-E_x(x_t^*)$, are bounded by the gap of approximation shown in \eqref{t14}. That is
\begin{align}
E_x(x_t^{s}) -\lambda m\bar D \leq F(x_t^s) \leq E_x(x_t^{s}),\label{a25}\\ 
E_x(x_t^*) -\lambda m\bar D \leq F(x_t^*) \leq E_x(x_t^*).
\label{a26}
\end{align}
That is, $E_x(x_t^{s})-F(x_t^s)$ is upper-bounded by $\lambda m\bar D$, and $F(x_t^*)-E_x(x_t^*)$ is upper-bounded by 0.
From Theorem \ref{th1}, the second pair $F(x_t^s)-F(x_t^*)$ is bounded by \eqref{t11}. Therefore, we can conclude \eqref{t21}. This completes the proof.

\paragraph{Proof of Theorem \ref{t3}}
It is seen from \eqref{c11} that $u_t^{j+1}>0$ given $u_t^{j}>0$ with a normalized matrix $B$. Also, we observe a trade-off between effects on the update from $|u_t^j|$ and $e^{-Bu_t^j}$, either one deviating zero while the other approaching zero. Based on the fact that $\lim_{u_t^j\rightarrow 0} u_t^j.(\bar Be^{-Bu_t^j}) =0$ and $\lim_{u_t^j\rightarrow \infty} u_t^j.(\bar B e^{-Bu_t^j}) =0$ where $\bar B$ is a constant matrix with non-negative elements, we can infer that the update \eqref{c11} will not diverge but will have an upper bound for the updated $u^{j+1}$. Recalling Algorithm \ref{al3} and the condition \eqref{c10}, we can write \eqref{c11} as
\begin{align}
u_t^{j+1}-u_t^j&=R_{u}^j(-\nabla f_u(u_t^{j}))-u_t^{j}\nonumber\\
&=-R_{u}^j(\nabla f_u(u_t^{j})+(R_{u}^j)^{-1}u_t^{j})\nonumber\\
&=-R_{u}^j(\nabla f_u(u_t^{j})+\nabla_{u_t} h_u(u_t^{j},u_t^{j}))\nonumber\\
&=-R_{u}^j\nabla_{u_t} H_u(u_t^{j}, u_t^{j})
\label{a33}
\end{align}
It follows from \eqref{c12} that $R_{u}^j> 0$ is a diagonal matrix during the learning. Therefore, the update law in Algorithm \ref{al3} for causes admits a gradient descent form with a positive-definite diagonal matrix as step size during the learning. The learning will stop when $R_{u}^j=0$, i.e., $u^j=0$, and $H_u(u_t^{j}, .)$ is minimized. That is, the method will learn until $u_t$ becomes sparse and the optimal condition \eqref{c10} is met. By taking the first two orders of Taylor expansion of $H_u(u_t^{j+1}, .)$, we have
\begin{align}
H_u(u_t^{j+1}, u_t^j)&= H_u(u_t^{j}, u_t^j)+(\nabla_{u_t} H_u(u_t^{j}, u_t^j))^T(u_t^{j+1}-u_t^{j})\nonumber\\
&=H_u(u_t^{j}, u_t^j)-(\nabla_{u_t} H_u(u_t^{j}, u_t^j))^TR_u^j(\nabla_{u_t} H_u(u_t^{j}, u_t^j))\nonumber\\
&\leq H_u(u_t^{j}, u_t^j)
\label{a34}
\end{align}
Combining it with \eqref{a29}-\eqref{a30} yields
\begin{align}
E_u(u_t^{j+1}) &=H_u(u_t^{j+1},u_t^{j+1})\leq H_u(u_t^{j+1},u_t^j)\leq H_u(u_t^{j},u_t^j)=E_u(u_t^{j})
\label{a35}
\end{align}
with equality at $u_t^{j+1}=u_t^{j}$. It can be concluded that function $E_u$ decreases using Algorithm \ref{al3} for causes inference. This convergence is also verified by the experiments.

Lemma \ref{lm1} still holds if we replace $f(x_t), h(x_t,V), F, H_x$ with $g(u_t), h(u_t,V), E_u, H_u$, respectively. Let $V=u_t^j$ and $P(V)=u_t^{j+1}$, and let $u_t^*$ be the optimal solution satisfying \eqref{c10}. Following Theorem \ref{th1} we have
\begin{align}
    E_u(u_t^s)-E_u(u_t^*)&\leq  \frac{1}{2s}\sum_{j=0}^{s-1} \sum_{k=1}^D \frac{(|(u_t^*)_k|-|(u_t^{j})_k|)^2 }{|(u_t^{j})_k|},
\end{align}
and consequently \eqref{a32}. This completes the proof.

\section{Appendix for more results and AE architecture details}
\label{res-AE-arch}
We used a simple geometric moving shape data set to demonstrate the video clustering mechanism for MM-DPCN further. Each video contains three geometric shapes: diamond, triangle, and square. Each shape appears consistently for 100 frames until another shape shows up. The shape could appear in each patch of the image frame and move within the 100 frames of a single shape.

\begin{figure}[t]
    \centering
    \includegraphics[width=\linewidth]{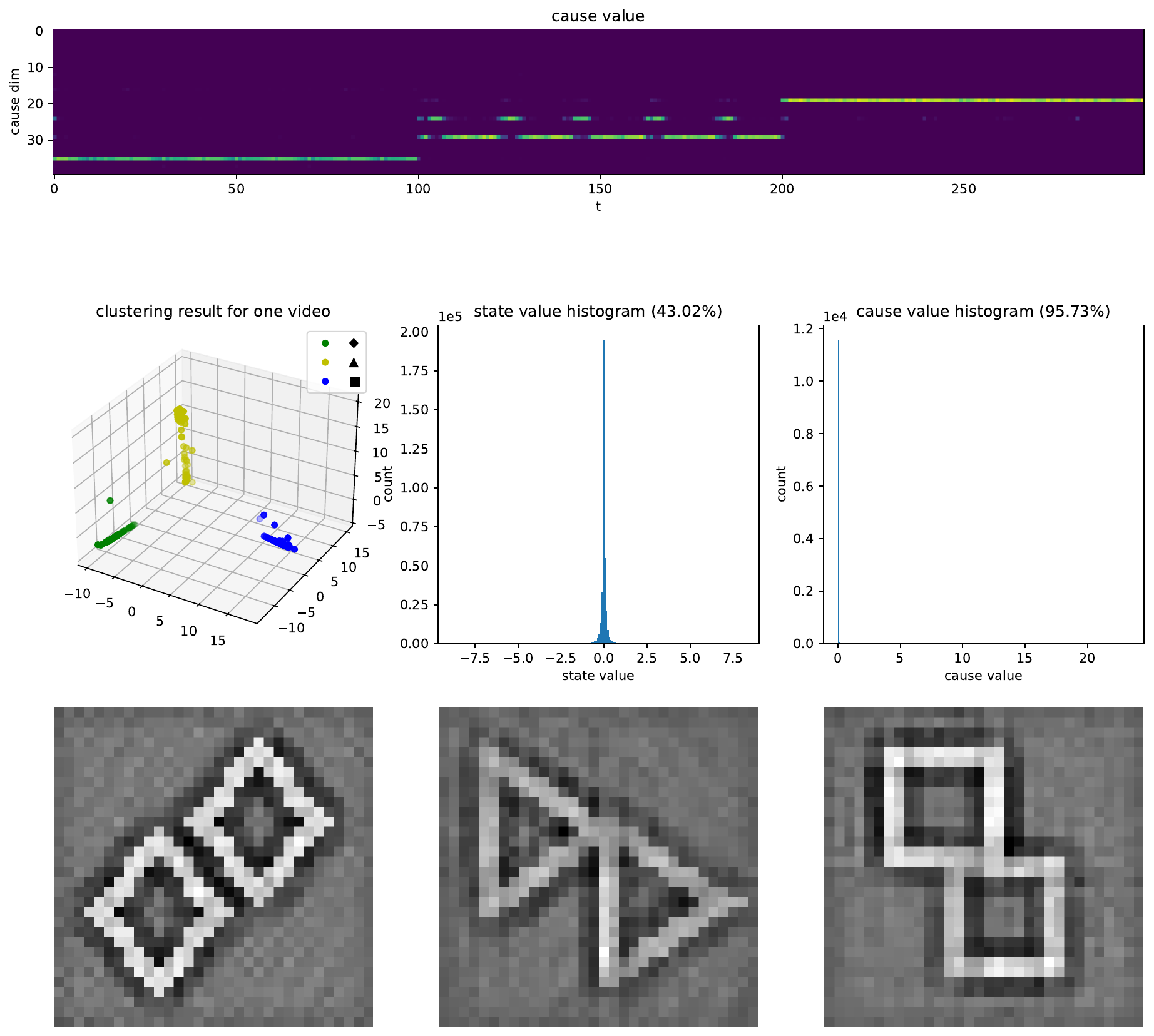}
    \caption{Clustering results for the moving geometric shape data set. The first row is the cause vector plot for one video. The three shapes are perfectly orthogonalized and assigned to the correct clusters. The third row shows examples of reconstruction.}
    \label{shape-clustering}
\end{figure}
\begin{figure}[ht]
    \centering
    \includegraphics[width=\linewidth]{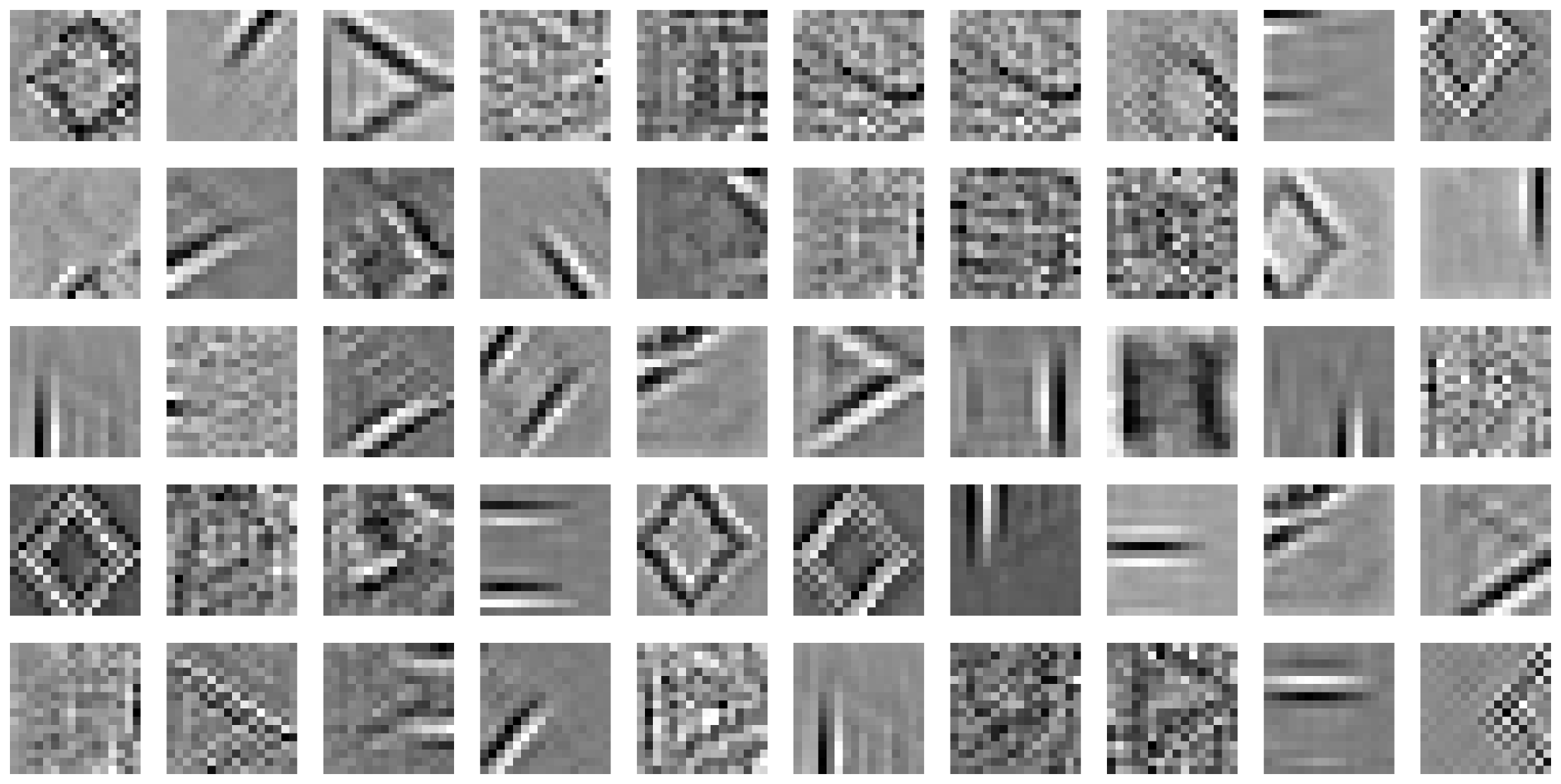}
    \caption{Learned filters $C^1$ on moving geometric shape data set.}
    \label{filter-shape}
\end{figure}

To visualize the learned filters, the plots for matrix $C^1$ are provided in Fig.~\ref{filter-shape}. 
The architectures for AE and WTA-RNN-AE used for the comparison results are provided in Table \ref{AE-arch}. We use the same architectures for both Mario and Coil-100 data sets.
\begin{table}[ht]
    \centering
    \caption{AE and WTA-RNN-AE architectures.}
    \label{AE-arch}
    \begin{tabular}{ccc}
    \toprule
        layer name & AE & WTA-RNN-AE \\
        \midrule
        encoder\_layer1 & $\left[3, 128\right]$ & $\left[3, 256\right]$\\
        encoder\_layer2 & $\left[128, 64\right]$ & $\left[256, 128\right]$\\
        encoder\_layer3 & $\left[64, 36\right]$ & $\left[128, 64\right]$\\
        encoder\_layer4 & $\left[36, 18\right]$ & *\\
        encoder\_layer5 & $\left[18, 9\right]$ & *\\
        RNN & * & $\left[64, 64\right]\times 5$\\
        \bottomrule
    \end{tabular}
\end{table}

\end{document}